\documentclass[11pt,a4paper]{article}
\usepackage[hyperref]{acl2017}
\usepackage{times}
\usepackage{latexsym}
\usepackage{amsmath, amssymb, amsthm, enumerate, framed, graphicx}
\usepackage{url}
\usepackage{booktabs}
\usepackage{multirow}
\usepackage{subfig}
\usepackage{pifont}
\usepackage{tikz}
\usepackage{latexsym}
\usepackage{color}
\usepackage{enumitem}
\usepackage[ruled,noresetcount,linesnumbered]{algorithm2e}
\usepackage{scalerel}
\usepackage{booktabs}
\usepackage{pbox}
\usepackage{todonotes}
\usepackage[font=small]{caption}
\usepackage{listings}
\usepackage[export]{adjustbox}
\usetikzlibrary{tikzmark}

\DeclareFixedFont{\ttb}{T1}{txtt}{bx}{n}{12} 
\DeclareFixedFont{\ttm}{T1}{txtt}{m}{n}{12}  

\usepackage{color}
\definecolor{deepblue}{rgb}{0,0,0.5}
\definecolor{deepred}{rgb}{0.6,0,0}
\definecolor{deepgreen}{rgb}{0,0.5,0}

\definecolor{fillcolor}{RGB}{216,217,252}

\renewcommand{\tt}[1]{\fontfamily{cmtt}\selectfont #1}

\DeclareMathOperator*{\argmax}{arg\,max}

\newcommand{\tdag}[1]{\ensuremath{{#1}^{\dag}}}
\newcommand{\tddag}[1]{\ensuremath{{#1}^{\ddag}}}
\usepackage{xcolor}
\aclfinalcopy 

\newcommand{\xsv}{\mathbf{x}^{(s)}}
\newcommand{\xtv}{\mathbf{x}^{(t)}}
\newcommand{\xtvhat}{\hat{\mathbf{x}}^{(t)}}

\newcommand{\ytv}{\mathbf{y}^{(t)}}

\newcommand{\yt}{y^{(t)}}
\newcommand{\xv}{\mathbf{x}}
\newcommand{\yv}{\mathbf{y}}
\newcommand{\zv}{\mathbf{z}}
\newcommand{\eps}{\mathbf{\epsilon}}

\title{Multi-space Variational Encoder-Decoders \\ for Semi-supervised Labeled Sequence Transduction
}
\author{Chunting Zhou, Graham Neubig \\
  Language Technologies Institute \\
  Carnegie Mellon University \\
  {\tt {ctzhou,gneubig}@cs.cmu.edu} \\}

\date{}

\begin{document}
\maketitle
\begin{abstract}
Labeled sequence transduction is a task of transforming one sequence into another sequence that satisfies desiderata specified by a set of labels. In this paper we propose \textit{multi-space variational encoder-decoders}, a new model for labeled sequence transduction with semi-supervised learning. The generative model can use neural networks to handle both discrete and continuous latent variables to exploit various features of data. Experiments show that our model provides not only a powerful supervised framework but also can effectively take advantage of the unlabeled data. On the SIGMORPHON morphological inflection benchmark, our model outperforms single-model state-of-art results by a large margin for the majority of languages.%
\footnote{An implementation of our model are available at \url{https://github.com/violet-zct/MSVED-morph-reinflection}.}
\end{abstract}
\section{Introduction}

This paper proposes a model for labeled sequence transduction tasks, tasks where we are given an input sequence and a set of labels, from which we are expected to generate an output sequence that reflects the content of the input sequence and desiderata specified by the labels.
Several examples of these tasks exist in prior work: using labels to moderate politeness in machine translation results \cite{sennrich2016controlling}, modifying the output language of a machine translation system \cite{johnson2016google}, or controlling the length of a summary in summarization \cite{kikuchi2016outputlength}.
In particular, however, we are motivated by the task of morphological reinflection \cite{sigmorphon}, which we will use as an example in our description and test bed for our models. 

\begin{figure}[tb]
  \centering
  \includegraphics[scale=0.35]{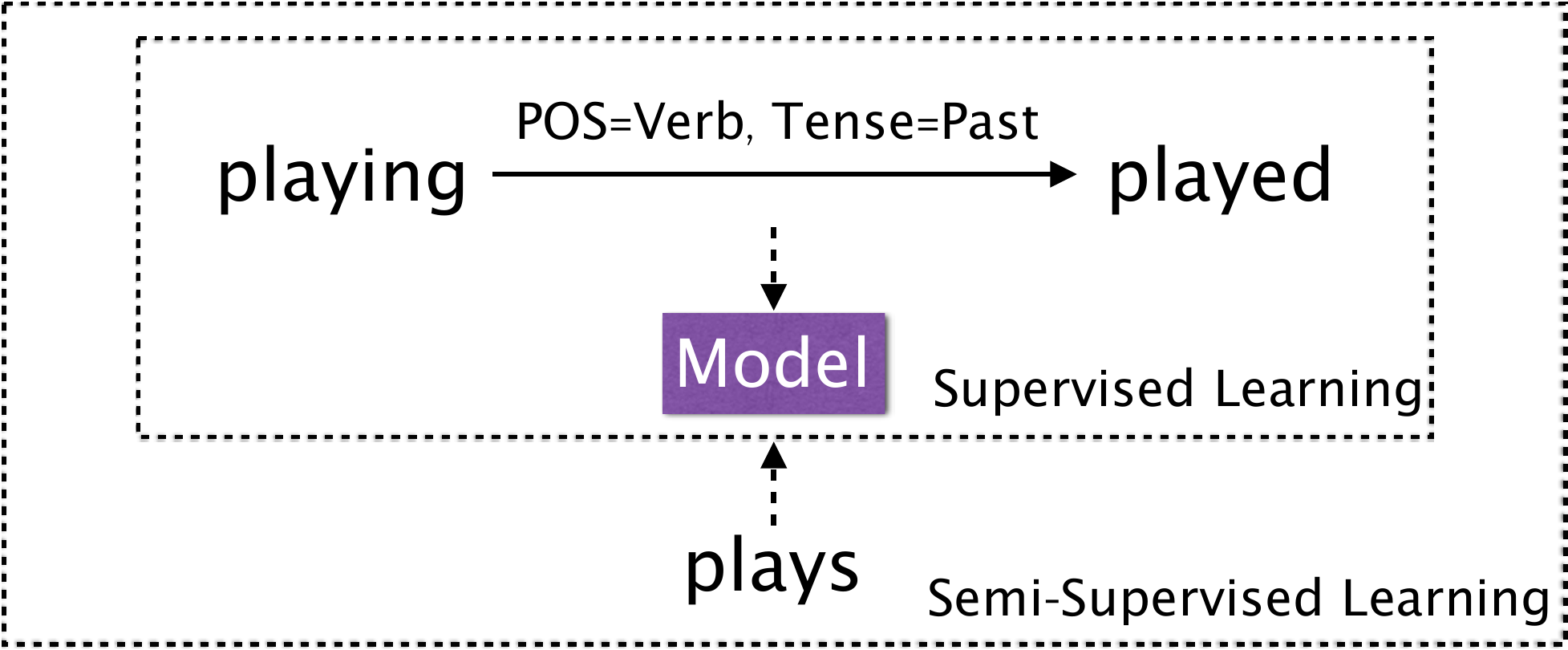}
  \caption{Standard supervised labeled sequence transduction, and our proposed semi-supervised method.}
  \label{fig:examples}
  \vspace{-3mm}
\end{figure}

In morphologically rich languages, different affixes (i.e. prefixes, infixes, suffixes) can be combined with the lemma to reflect various syntactic and semantic features of a word.
The ability to accurately analyze and generate morphological forms is crucial to creating applications such as machine translation \cite{chahuneau2013translating,toutanova2008applying} or information retrieval \cite{darwish2007adapting} in these languages.
As shown in \ref{fig:examples}, \textit{re}-inflection of an inflected form given the target linguistic labels is a challenging sub-task of handling morphology as a whole, in which we take as input an inflected form (in the example, ``playing'') and labels representing the desired form (``{\tt pos=Verb, tense=Past}'') and must generate the desired form (``played'').

Approaches to this task include those utilizing hand-crafted linguistic rules and heuristics \cite{taji}, as well as learning-based approaches using alignment and extracted transduction rules \cite{durrett2013supervisedmorphology,ehu,alberta}.
There have also been methods proposed using neural sequence-to-sequence models \cite{faruqui2016morphological,kann2016neural,ostling2016morphological}, and currently ensembles of attentional encoder-decoder models~\citep{med,med2} have achieved state-of-art results on this task. 
One feature of these neural models however, is that they are trained in a largely \textit{supervised} fashion (top of Fig.~\ref{fig:examples}), using data explicitly labeled with the input sequence and labels, along with the output representation.
Needless to say, the ability to obtain this annotated data for many languages is limited.
However, we can expect that for most languages we can obtain large amounts of unlabeled surface forms that may allow for \textit{semi-supervised} learning over this unlabeled data (entirety of Fig.~\ref{fig:examples}).\footnote{\newcite{faruqui2016morphological} have attempted a limited form of semi-supervised learning by re-ranking with a standard $n$-gram language model, but this is not integrated with the learning process for the neural model and gains are limited.}

In this work, we propose a new framework for labeled sequence transduction problems: multi-space variational encoder-decoders ({\bf MSVED}, \S\ref{sec:msved}).
MSVEDs employ continuous or discrete latent variables belonging to multiple separate probability distributions\footnote{Analogous to multi-space hidden Markov models \cite{tokuda2002multi}} to explain the observed data.
In the example of morphological reinflection, we introduce a vector of continuous random variables that represent the lemma of the source and target words, and also one discrete random variable for each of the labels, which are on the source or the target side.

This model has the advantage of both providing a powerful modeling framework for supervised learning, and allowing for learning in an unsupervised setting.
For labeled data, we maximize the variational lower bound on the marginal log likelihood of the data and annotated labels.
For unlabeled data, we train an auto-encoder to reconstruct a word conditioned on its lemma and morphological labels.
While these labels are unavailable, a set of discrete latent variables are associated with each unlabeled word.
Afterwards we can perform posterior inference on these latent variables and maximize the variational lower bound on the marginal log likelihood of data. 

Experiments on the SIGMORPHON morphological reinflection task \cite{sigmorphon} find that our model beats the state-of-the-art for a single model in the majority of languages, and is particularly effective in languages with more complicated inflectional phenomena.
Further, we find that semi-supervised learning allows for significant further gains.
Finally, qualitative evaluation of lemma representations finds that our model is able to learn lemma embeddings that match with human intuition.

\section{Labeled Sequence Transduction}
\label{sec:problem}

In this section, we first present some notations regarding labeled sequence transduction problems in general, then describe a particular instantiation for morphological reinflection.

\noindent{\bf Notation: }
Labeled sequence transduction problems involve transforming a source sequence $\xsv$ into a target sequence $\xtv$, with some labels describing the particular variety of transformation to be performed.
We use discrete variables $\yt_1, \yt_2, \cdots, \yt_K$ to denote the labels associated with each target sequence, where $K$ is the total number of labels. Let $\ytv=[\yt_1, \yt_2, \cdots, \yt_K]$ denote a vector of these discrete variables. Each discrete variable $\yt_k$ represents a categorical feature pertaining to the target sequence, and has a set of possible labels. In the later sections, we also use $\ytv$ and $\yt_k$ to denote discrete latent variables corresponding to these labels.

Given a source sequence $\xsv$ and a set of associated target labels $\ytv$, 
our goal is to generate a target sequence $\xtv$ that exhibits the features specified by $\ytv$ using a probabilistic model $p(\xtv | \xsv, \ytv)$. 
The best target sequence $\xtvhat$ is then given by:
\begin{equation}
\xtvhat = \argmax_{\xtv} p(\xtv | \xsv, \ytv).
\end{equation}

\noindent{\bf Morphological Reinflection Problem: }
In morphological reinflection, the source sequence $\xsv$ consists of the characters in an inflected word (e.g., ``{\tt played}''), while the associated labels $\ytv$ describe some linguistic features (e.g., $\yt_\textrm{pos}=\texttt{Verb}$, $\yt_\textrm{tense}=\texttt{Past}$) that we hope to realize in the target. The target sequence $\xtv$ is therefore the characters of the re-inflected form of the source word (e.g., ``{\tt played}'') that satisfy the linguistic features specified by $\ytv$. For this task, each discrete variable $\yt_k$ has a set of possible labels (e.g. {\tt pos=V, pos=ADJ}, etc) and follows a multinomial distribution.


\section{Proposed Method}
\label{sec:proposed}
\subsection{Preliminaries: Variational Autoencoder}
\begin{figure}[tb]
  \centering
  \includegraphics[scale=0.25]{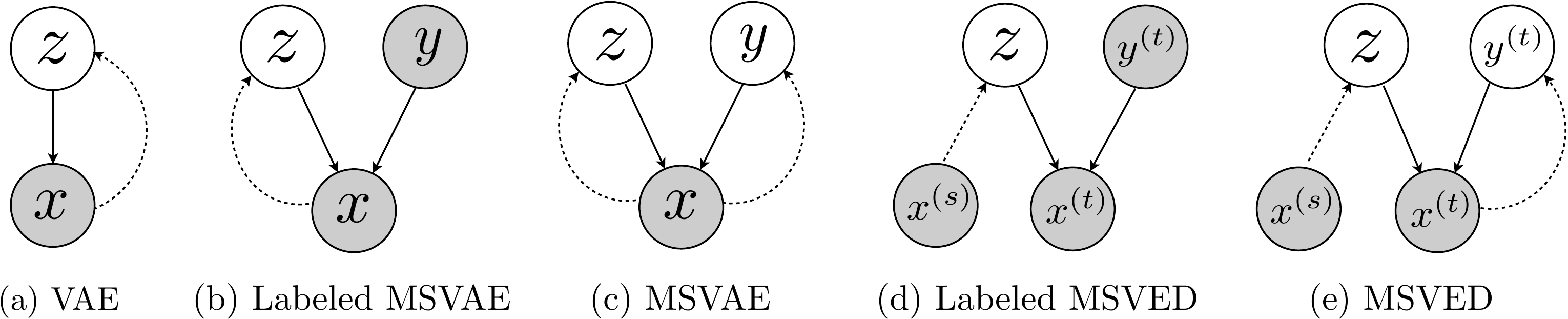}
  \caption{Graphical models of (a) VAE, (b) labeled MSVAE, (c) MSVAE, (d) labeled MSVED, and (e) MSVED. White circles are latent variables and shaded circles are observed variables.
Dashed lines indicate the inference process while the solid lines indicate the generative process.}
  \label{fig:pgm}
  \vspace{-2mm}
\end{figure}

As mentioned above, our proposed model uses probabilistic latent variables in a model based on neural networks.
The variational autoencoder \cite{vae} is an efficient way to handle (continuous) latent variables in neural models.
We describe it briefly here, and interested readers can refer to \newcite{doersch2016tutorial} for details.
The VAE learns a generative model of the probability $p(\xv|\zv)$ of observed data $\xv$ given a latent variable $\zv$, and simultaneously uses a recognition model $q(\zv|\xv)$ at learning time to estimate $\zv$ for a particular observation $\xv$ (Fig.~\ref{fig:pgm}(a)).
$q(\cdot)$ and $p(\cdot)$ are modeled using neural networks parameterized by $\phi$ and $\theta$ respectively, and these parameters are learned by maximizing the variational lower bound on the marginal log likelihood of data:
\begin{align}
    \nonumber\log p_{\theta}(\xv) &\geq
     \mathbb{E}_{\zv\sim q_{\phi}(\zv|\xv)}[\log p_{\theta}(\xv|\zv)] -\\ &~~~~\text{KL}(q_{\phi}(\zv|\xv)||p(\zv))
    \label{vae}
\end{align}
The KL-divergence term (a standard feature of variational methods) ensures that the distributions estimated by the recognition model $q_{\phi}(\zv|\xv)$ do not deviate far from our prior probability $p(\zv)$ of the values of the latent variables.
To optimize the parameters with gradient descent, \newcite{vae} introduce a reparameterization trick that allows for training using simple backpropagation w.r.t.~the Gaussian latent variables $\zv$. Specifically, we can express $\zv$ as a deterministic variable $\zv = g_{\phi}(\eps, \xv)$ where $\eps$ is an independent Gaussian noise variable $\eps \sim \mathcal{N}(0, 1)$.  The mean $\mu$ and the variance $\sigma^2$ of $\zv$ are reparameterized by the differentiable functions w.r.t. $\phi$. Thus, instead of generating $\zv$ from $q_{\phi}(\zv|\xv)$, we sample the auxiliary variable $\eps$ and obtain $\zv = \mu_{\phi}(x) + \sigma_{\phi}(x) \circ \eps$, which enables gradients to backpropagate through $\phi$.

\subsection{Multi-space Variational Autoencoders}
As an intermediate step to our full model, we next describe a generative model for a single sequence with both continuous and discrete latent variables, the multi-space variational auto-encoder (MSVAE).
MSVAEs are a combination of two threads of previous work: deep generative models with both continuous/discrete latent variables for classification problems \cite{semivae,maaloe2016auxiliary} and VAEs with only continuous variables for sequential data \cite{bowman,vrnn,vnmt,fabius2014variational,bayer2014learning}.
In MSVAEs, we have an observed sequence $\xv$, continuous latent variables $\zv$ like the VAE, as well as discrete variables $\yv$.

In the case of the morphology example, $\xv$ can be interpreted as an inflected word to be generated.
$\yv$ is a vector representing its linguistic labels, either annotated by an annotator in the observed case, or unannotated in the unobserved case.
$\zv$ is a vector of latent continuous variables, e.g. a latent embedding of the lemma that captures all the information about $\xv$ that is not already represented in labels $\yv$.

\noindent\textbf{MSVAE}: 
Because inflected words can be naturally thought of as ``lemma+morphological labels'', to interpret a word, we resort to discrete and continuous latent variables that represent the linguistic labels and the lemma respectively. In this case when the labels of the sequence $\yv$ is not observed, we perform inference over possible linguistic labels and these inferred labels are referenced in generating $\xv$.

The generative model $p_{\theta}(\xv, \yv, \zv) = p(\zv)p_{\mathbf{\pi}}(\yv)p_{\theta}(\xv|\yv,\zv)$ is defined as:
\begin{align}
p(\zv) &= \mathcal{N}(\zv|\mathbf{0}, \mathbf{I}) \label{equ:zprior}\\
p_{\mathbf{\pi}}(\yv) &= \prod_k\text{Cat}(y_k|\mathbf{\pi}_k) \label{equ:yprior}\\
p_{\theta}(\xv|\yv, \zv) &= f(\xv; \yv, \zv, \theta) \label{equ:msvae}.
\end{align}
Like the standard VAE, we assume the prior of the latent variable $\zv$ is a diagonal Gaussian distribution with zero mean and unit variance.
We assume that each variable in $\yv$ is independent, resulting in a factorized distribution in Eq. \ref{equ:yprior}, where $\text{Cat}(y_k|\mathbf{\pi}_k)$ is a multinomial distribution with parameters $\mathbf{\pi}_k$.
For the purposes of this study, we set these to a uniform distribution $\pi_{k,j}=\frac{1}{|\mathbf{\pi}_k|}$.
$f(\xv; \yv, \zv, \theta)$ calculates the likelihood of $\xv$, a function parametrized by deep neural networks.
Specifically, we employ an RNN decoder to generate the target word conditioned on the lemma variable $\zv$ and linguistic labels $\yv$, detailed in \S\ref{sec:neuralmodel}.

When inferring the latent variables from the given data $\xv$, we assume the joint distribution of latent variables $\zv$ and $\yv$ has a factorized form, i.e. $q(\zv, \yv|\xv) = q(\zv|\xv)q(\yv|\xv)$ as shown in Fig.~\ref{fig:pgm}(c).
The inference model is defined as follows:
\begin{align}
    q_{\phi}(\zv|\xv) &= \mathcal{N}(\mathbf{z|\mu_{\phi}(x)}, \text{diag}(\sigma^2_{\phi}(\xv))) \label{infer}\\
  \nonumber q_{\phi}(\yv|\xv) &= \prod_k q_{\phi}(y_k|\xv) \\
   &= \prod_k\text{Cat}(y_k|\mathbf{\pi_{\phi}(x)})
\end{align}
where the inference distribution over $\zv$ is a diagonal Gaussian distribution with mean and variance parameterized by neural networks.
The inference model $q(\yv|\xv)$ on labels $\yv$ has the form of a discriminative classifier that generates a set of multinomial probability vectors $\mathbf{\pi_{\phi}(x)}$ over all labels for each tag $y_k$.
We represent each multinomial distribution $q(y_k|\xv)$ with an MLP.

The MSVAE is trained by maximizing the following variational lower bound $\mathcal{U}(\xv)$ on the objective for unlabeled data:
\begin{align}
\nonumber\log p_{\theta}(\xv) &\geq \mathbb{E}_{(\yv, \zv)\sim q_{\phi}(\mathbf{y,z|x})}\log\frac{p_{\theta}(\xv, \yv, \zv)}{q_{\phi}(\yv,\zv|\xv
)}\\&\nonumber=\mathbb{E}_{\yv\sim q_{\phi}(\yv|\xv)}[\mathbb{E}_{\zv\sim q_{\phi}(\zv|\xv)}[\log p_{\theta}(\xv|\zv, \yv)] \\\nonumber &- \text{KL}(q_{\phi}(\zv|\xv)||p(\zv)) + \log p_{\mathbf{\pi}}(\yv) \\ &- \log q_{\phi}(\yv|\xv)] = \mathcal{U}(\xv)
\label{hybrid-vae}
\end{align}
Note that this introduction of discrete variables requires more sophisticated optimization algorithms, which we will discuss in \S\ref{sec:gumbel}.

\noindent\textbf{Labeled MSVAE}: 
When $\yv$ is observed as shown in Fig.~\ref{fig:pgm}(b), we maximize the following variational lower bound on the marginal log likelihood of the data and the labels:
\begin{align}
\nonumber\log p_{\theta}(\xv, \yv) &\geq \mathbb{E}_{\zv\sim q_{\phi}(\zv|\xv)}\log\frac{p_{\theta}(\xv, \yv, \zv)}{q_{\phi}(\zv|\xv)}=\\ 
\nonumber& \mathbb{E}_{\zv\sim q_{\phi}(\zv|\xv)}[\log p_{\theta}(\xv|\yv,\zv) + \log p_{\mathbf{\pi}}(\yv)] \\&~~~~~- \text{KL}(q_{\phi}(\zv|\xv)||p(\zv))
\end{align}
which is a simple extension to Eq. \ref{vae}.

Note that when labels are not observed, the inference model $q_{\phi}(\yv|\xv)$ has the form of a discriminative classifier, thus we can use observed labels as the supervision signal to learn a better classifier.
In this case we also minimize the following cross entropy as the classification loss:
\begin{equation}
    \mathcal{D}(\xv, \yv) = \mathbb{E}_{(\xv, \yv)\sim p_{l}(\xv, \yv)}[-\log q_{\phi}(\yv|\xv)]
    \label{eq:err}
\end{equation}
where $p_{l}(\xv, \yv)$ is the distribution of labeled data.
This is a form of multi-task learning, as this additional loss also informs the learning of our representations.

\subsection{Multi-space Variational Encoder-Decoders} 
\label{sec:msved}
Finally, we discuss the full proposed method: the multi-space variational encoder-decoder (MSVED), which generates the target $\xtv$ from the source $\xsv$ and labels $\ytv$. Again, we discuss two cases of this model: labels of the target sequence are observed and not observed.

\noindent\textbf{MSVED:} 
The graphical model for the MSVED is given in Fig.~\ref{fig:pgm} (e).
Because the labels of target sequence are not observed, once again we treat them as discrete latent variables and make inference on the these labels conditioned on the target sequence.
The generative process for the MSVED is very similar to that of the MSVAE with one important exception: while the standard MSVAE conditions the recognition model $q(\zv|\xv)$ on $\xv$, then generates $\xv$ itself, the MSVED conditions the recognition model $q(\zv|\xsv)$ on the source $\xsv$, then generates the target $\xtv$.
Because only the recognition model is changed, the generative equations for $p_{\theta}(\xtv, \ytv, \zv)$ are exactly the same as Eqs.~\ref{equ:zprior}--\ref{equ:msvae} with $\xtv$ swapped for $\xv$ and $\ytv$ swapped for $\yv$. 
The variational lower bound on the conditional log likelihood, however, is affected by the recognition model, and thus is computed as:
\begin{align}
    &\nonumber \log p_{\theta}(\xtv | \xsv) \\ \geq &\nonumber \mathbb{E}_{(\ytv, \zv)\sim q_{\phi}(\ytv, \zv|\xsv, \xtv)}\log\frac{p_{\theta}(\xtv, \ytv, \zv|\xsv)}{q_{\phi}(\ytv, \zv|\xsv, \xtv)} \\=
    &\nonumber \mathbb{E}_{\ytv\sim q_{\phi}(\ytv|\xtv)}[\mathbb{E}_{\zv\sim q_{\phi}(\zv|\xsv)}[\log p_{\theta}(\xtv|\ytv, \zv)] \\&- \nonumber\text{KL}(q_{\phi}(\zv|\xsv)||p(\zv))+ \log p_{\mathbf{\pi}}(\ytv) \\ &- \log q_{\phi}(\ytv|\xtv)] = \mathcal{L}_u(\xtv | \xsv)
\label{eq:unsup}
\end{align}
\noindent\textbf{Labeled MSVED:} When the complete form of $\xsv$, $\ytv$, and $\xtv$ is observed in our training data, the graphical model of the labeled MSVED model is illustrated in Fig.~\ref{fig:pgm} (d).
We maximize the variational lower bound on the conditional log likelihood of observing $\xtv$ and $\ytv$ as follows:
\begin{align}
    &~~~~\nonumber \log p_{\theta}(\xtv, \ytv | \xsv) \\\nonumber&\geq \mathbb{E}_{\zv\sim q_{\phi}(\zv|\xsv)}\log\frac{p_{\theta}(\xtv, \ytv, \zv|\xsv)}{q_{\phi}(\zv|\xsv)} \\
    \nonumber&= \mathbb{E}_{\zv\sim q_{\phi}(\zv|\xsv)}[\log p_{\theta}(\xtv|\ytv, \zv) + \log p_{\mathbf{\pi}}(\ytv)] -\\&\text{KL}(q_{\phi}(\zv|\xsv)||p(\zv)) = \mathcal{L}_l(\xtv, \ytv | \xsv)
\label{eq:sup}
\end{align}

\section{Learning MSVED}
\label{sec:learning}
Now that we have described our overall model, we discuss details of the learning process that prove useful to its success.
\subsection{Learning Discrete Latent Variables}
\label{sec:gumbel}
One challenge in training our model is that it is not trivial to perform back-propagation through discrete random variables, and thus it is difficult to learn in the models containing discrete tags such as MSVAE or MSVED.\footnote{
 \newcite{semivae} solve this problem by marginalizing over all labels, but this is infeasible in our case where we have an exponential number of label combinations.
}
To alleviate this problem, we use the recently proposed Gumbel-Softmax trick \cite{maddison2014sampling,gumbel1954statistical,gumbel,gumbel2} to create a differentiable estimator for categorical variables.

The Gumbel-Max trick \cite{gumbel1954statistical} offers a simple way to draw samples from a categorical distribution
 with class probabilities $\pi_{1}, \pi_{2}, \cdots$ by using the {\tt argmax}  operation as follows: $\text{one\_hot}(\argmax_i[g_i + \log \pi_i])$, where $g_{1}, g_{2}, \cdots$ are i.i.d. samples drawn from the Gumbel(0,1) distribution.%
 \footnote{The Gumbel (0,1) distribution can be sampled by first drawing $u\sim \text{Uniform(0,1)}$ and computing $g = -\log(-\log(u))$.}
When making inferences on the morphological labels $y_1, y_2, \cdots$, the Gumbel-Max trick can be approximated by the continuous softmax function with temperature $\tau$ to generate a sample vector $\mathbf{\hat{y}}_i$ for each label $i$:
\begin{align}
    \hat{y}_{ij} = \frac{\exp((\log(\pi_{ij}) + g_{ij})/\tau)}{\sum_{k=1}^{N_i}{\exp((\log(\pi_{ik}) + g_{ik})/\tau}}
\end{align}
where $N_i$ is the number of classes of label $i$. When $\tau$ approaches zero, the generated sample $\mathbf{\hat{y}}_i$ becomes a one-hot vector.
When $\tau > 0$, $\mathbf{\hat{y}}_i$ is smooth w.r.t $\mathbf{\pi}_i$.
In experiments, we start with a relatively large temperature and decrease it gradually.

\subsection{Learning Continuous Latent Variables}

MSVED aims at generating the target sequence conditioned on the latent variable $\zv$ and the target labels $\ytv$. This requires the encoder to generate an informative representation $\zv$ encoding the content of the $\xsv$. However, the variational lower bound in our loss function contains the KL-divergence between the approximate posterior $q_{\phi}(\zv|\xv)$ and the prior $p(\zv)$, which is relatively easy to learn compared with learning to generate output from a latent representation. We observe that with the vanilla implementation the KL cost quickly decreases to near zero, setting $q_{\phi}(\zv|\xv)$ equal to standard normal distribution. In this case, the RNN decoder can easily rely on the true output of last time step during training to decode the next token, which degenerates into an RNN language model. Hence, the latent variables are ignored by the decoder and cannot encode any useful information. The latent variable $\zv$ learns an undesirable distribution that coincides with the imposed prior distribution but has no contribution to the decoder. To force the decoder to use the latent variables, we take the following two approaches which are similar to \newcite{bowman}.

\noindent\textbf{KL-Divergence Annealing: }We add a coefficient $\mathbf{\lambda}$ to the KL cost and gradually anneal it from zero to a predefined threshold $\mathbf{\lambda_m}$.  At the early stage of training,
we set $\lambda$ to be zero and let the model first figure out how to project the representation of the source sequence to a roughly right point in the space and then regularize it with the KL cost. Although we are not optimizing the tight variational lower bound, the model balances well between generation and regularization. This technique can also be seen in \cite{kovcisky2016semantic,miao2016language}.\\
\textbf{Input Dropout in the Decoder:} Besides annealing the KL cost, we also randomly drop out the input token with a probability of $\beta$ at each time step of the decoder during learning.  The previous ground-truth token embedding is replaced with a zero vector when dropped. In this way, the RNN decoder could not fully rely on the ground-truth previous token, which ensures that the decoder uses information encoded in the latent variables.

\begin{table*}[ht!]
\centering
\small
\begin{tabular}{lrr|lll|lc}
\toprule \
                  &               &                & \multicolumn{3}{c|}{\textbf{Proposed MSVED}} & \multicolumn{2}{c}{\textbf{Baseline MED}} \\
\textbf{Language} & \textbf{\#LD} & \textbf{\#ULD} & \textbf{SD-Sup} & \textbf{BD-Sup} & \textbf{Semi-sup} & \textbf{Single} & \textbf{Ensemble} \\\midrule
Turkish           & 12,798         & 29,608	         & 93.25 & \tdag{95.66} & \tddag{\textbf{97.25}} & 89.56 & 95.00           \\
Hungarian         & 19,200         & 34,025          & 97.00 & \tdag{98.54} & \tddag{\textbf{99.16}} & 96.46 & 98.37          \\
Spanish           & 12,799         & 72,151	         & 88.32 & 91.50 & 93.74          & \tddag{\tdag{94.74}} & \textbf{96.69}  \\
Russian	          & 12,798         & 67,691          & 75.77 & 83.07 & \tddag{86.80}          & \tdag{83.55} & \textbf{87.13}  \\
Navajo            & 12,635         & 6,839           & 85.00 & \tdag{95.37} & \tddag{\textbf{98.25}} & 63.62 & 83.00           \\
Maltese           & 19,200         & 46,918	         & 84.83 & \tdag{88.21} & \tddag{\textbf{88.46}} & 79.59 & 84.25           \\
Arabic            & 12,797         & 53,791	         & 79.13 & \tdag{92.62} & \tddag{\textbf{93.37}} & 72.58 & 82.80           \\
Georgian          & 12,795         & 46,562          & 89.31 & \tdag{93.63} & \tddag{95.97}          & 91.06 & \textbf{96.21}  \\
German            & 12,777         & 56,246	         & 75.55 & 84.08 & \tddag{90.28}          & \tdag{89.11} & \textbf{92.41}  \\
Finnish           & 12,800         & 74,687	         & 75.59 & 85.11 & \tddag{91.20}          & \tdag{85.63} & \textbf{93.18}  \\
\midrule
Avg. Acc & -- & -- & 84.38 & \tdag{90.78} & \tddag{\textbf{93.45}} & 84.59 & 90.90\\ 
\bottomrule
\end{tabular}
\caption{Results for Task 3 of SIGMORPHON 2016 on Morphology Reinflection.
$\dag$ represents the best single supervised model score, $\ddag$ represents the best model including semi-supervised models, and bold represents the best score overall. {\bf \#LD} and {\bf \#ULD} are the number of supervised data and unlabeled words respectively.}
\label{tab:res}
\vspace{-4mm}
\end{table*}

\section{Architecture for Morphological Reinflection}
\label{sec:neuralmodel}
\textbf{Training details:}
For the morphological reinflection task, our supervised training data consists of source $\xsv$, target $\xtv$, and target tags $\ytv$.
We test three variants of our model trained using different types of data and different loss functions.
First, the single-directional supervised model ({\bf SD-Sup}) is purely supervised: it only decodes the target word from the given source word with the loss function $\mathcal{L}_l(\xtv, \ytv | \xsv)$ from Eq.~\ref{eq:sup}.
Second, the bi-directional supervised model ({\bf BD-Sup}) is trained in both directions: decoding the target word from the source word and decoding the source word from the target word, which corresponds to the loss function $\mathcal{L}_l(\xtv, \ytv | \xsv) + \mathcal{L}_u(\xsv | \xtv)$ using Eqs.~\ref{eq:unsup}--\ref{eq:sup}.
Finally, the semi-supervised model ({\bf Semi-sup}) is trained to maximize the variational lower bounds and minimize the classification cross-entropy error of \ref{eq:err}.
\begin{align}
\nonumber\mathcal{L}&(\mathbf{\xsv, \xtv, \ytv, x}) = \alpha\cdot\mathcal{U}(\xv) + \mathcal{L}_u(\xsv | \xtv) \\
      &+ \mathcal{L}_l(\xtv, \ytv | \xsv) - \mathcal{D}(\xtv, \ytv)
\label{fo}
\end{align}
The weight $\alpha$ controls the relative weight between the loss from unlabeled data and labeled data.

\begin{figure}[tb]
  \centering
  \includegraphics[scale=0.27]{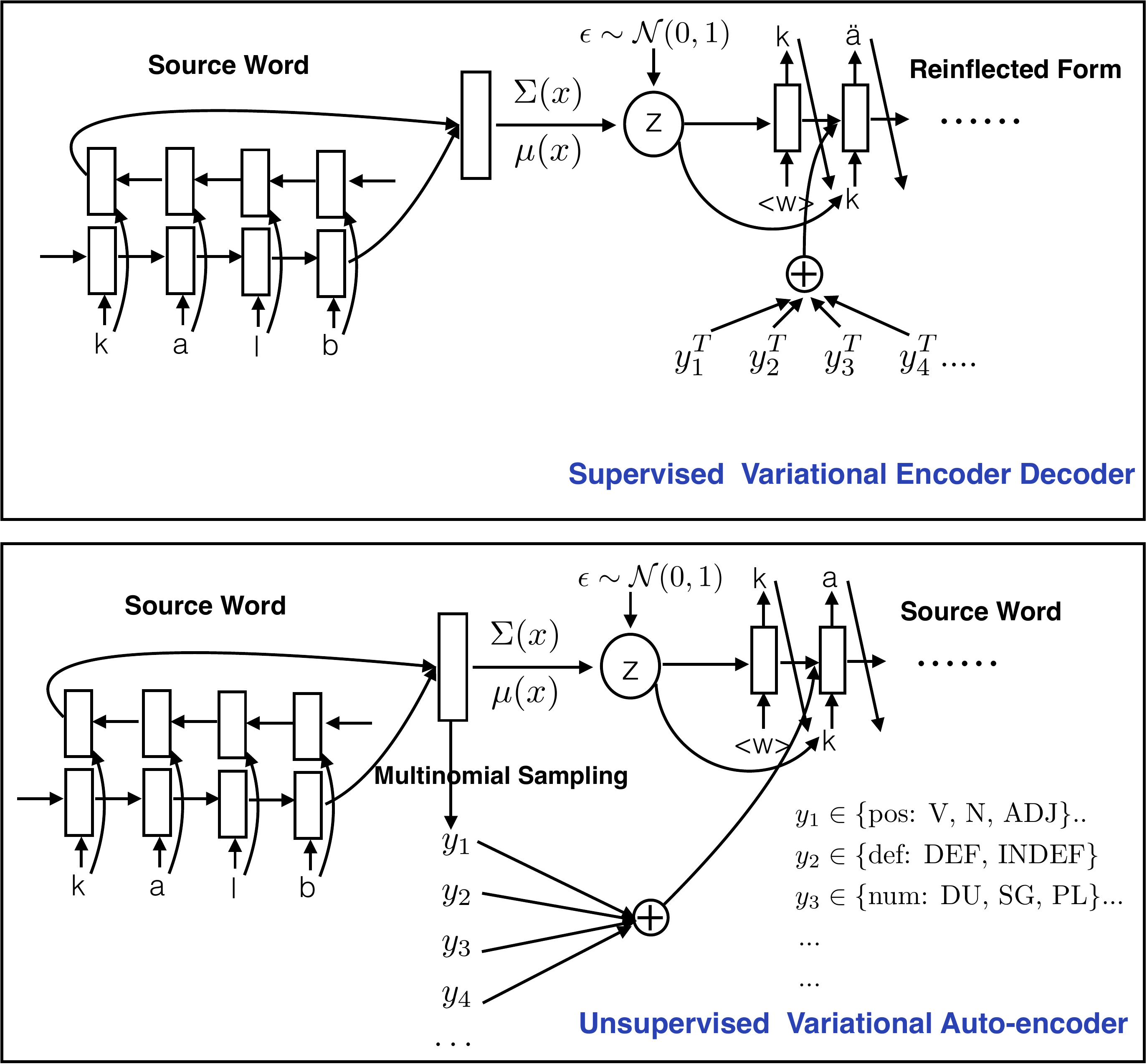}
  \caption{Model architecture for labeled and unlabeled data. For the encoder-decoder model, only one direction from the source to target is given. The classification model is not illustrated in the diagram.}
  \label{fig:sslvae}
  \vspace{-5mm}
\end{figure}

We use Monte Carlo methods to estimate the expectation over the posterior distribution $q(\zv|\xv)$ and $q(\yv|\xv)$ inside the objective function \ref{fo}. Specifically, we draw Gumbel noise and Gaussian noise one at a time to compute the latent variables $\yv$ and $\zv$.

The overall model architecture is shown in Fig.~\ref{fig:sslvae}.
Each character and each label is associated with a continuous vector.
We employ Gated Recurrent Units (GRUs) for the encoder and decoder.
Let $\mathbf{\overrightarrow{h_t}}$ and $\mathbf{\overleftarrow{h_t}}$ denote the hidden state of the forward and backward encoder RNN at time step $t$.
$\mathbf{u}$ is the hidden representation of $\xsv$ concatenating the last hidden state from both directions i.e. $[\mathbf{\overrightarrow{h_T}};\mathbf{\overleftarrow{h_T}}]$ where $T$ is the word length.
$\mathbf{u}$ is used as the input for the inference model on $\zv$.
We represent $\mathbf{\mu(u)}$ and $\mathbf{\sigma}^2\mathbf{(u)}$ as MLPs and sample $\zv$ from $\mathcal{N}(\mathbf{\mu(u)}, \text{diag}(\mathbf{\sigma}^2\mathbf{(u)}))$, using $\mathbf{z = \mu + \sigma \circ \epsilon}$, where $\epsilon\sim \mathcal{N}(\mathbf{0, I})$.
Similarly, we can obtain the hidden representation of $\xtv$ and use this as input to the inference model on each label $\ytv_i$ which is also an MLP following a softmax layer to generate the categorical probabilities of target labels.

In decoding, we use 3 types of information in calculating the probability of the next character :
(1) the current decoder state, (2) a tag context vector using attention \cite{bahdanau2014neural} over the tag embeddings, and (3) the latent variable $\zv$.
The intuition behind this design is that we would like the model to constantly consider the lemma represented by $\zv$, and also reference the tag corresponding to the current morpheme being generated at this point. We do not marginalize over the latent variable $\zv$ however, instead we use the mode $\mu$ of $\zv$ as the latent representation for $\zv$. We use beam search with a beam size of 8 to perform search over the character vocabulary at each decoding time step.

\noindent\textbf{Other experimental setups: }  
All hyperparameters are tuned on the validation set, and include the following:
For KL cost annealing, $\lambda_m$ is set to be 0.2 for all language settings.
For character drop-out at the decoder, we empirically set $\beta$ to be 0.4 for all languages.
We set the dimension of character embeddings to be 300, tag label embeddings to be 200, RNN hidden state to be 256, and latent variable $\zv$ to be 150.
We set $\alpha$ the weight for the unsupervised loss to be 0.8. 
We train the model with Adadelta \cite{zeiler2012adadelta} and use early-stop with a patience of 10.

\section{Experiments}
\subsection{Background: SIGMORPHON 2016}
SIGMORPHON 2016 is a shared task on morphological inflection over 10 different morphologically rich languages.
There are a total of three tasks, the most difficult of which is task 3, which requires the system to output the reinflection of an inflected word.%
\footnote{Task 1 is inflection of a lemma word and task 2 is reinflection but also provides the source word labels.}
The training data format in task 3 is in triples: (source word, target labels, target word). In the test phase, the system is asked to generate the target word given a source word and the target labels.
There are a total of three tracks for each task, divided based the amount of supervised data that can be used to solve the problem, among which track 2 has the strictest limitation of only using data for the corresponding task.
As this is an ideal testbed for our method, which can learn from unlabeled data, we choose track 2 and task 3 to test our our model's ability to exploit this data.

As a baseline, we compare our results with the MED system~\citep{med} which achieved state-of-the-art results in the shared task.
This system used an encoder-decoder model with attention on the concatenated source word and target labels.
Its best result is obtained from an ensemble of five RNN encoder-decoders ({\bf Ensemble}).
To make a fair comparison with our models, which don't use ensembling, we also calculated single model results ({\bf Single}).

All models are trained using the labeled training data provided for task 3.
For our semi-supervised model ({\bf Semi-sup}), we also leverage unlabeled data from the training and validation data for tasks 1 and 2 to train variational auto-encoders.
\begin{table}[ht!]
\centering
\small
\begin{tabular}{lccc}
\toprule \
\textbf{Language} & \textbf{Prefix} & \textbf{Stem} & \textbf{Suffix} \\\midrule
Turkish & 0.21 & 1.12 & 98.75\\
Hungarian & 0.00 & 0.08 & 99.79 \\
Spanish & 0.09 & 3.25 & 90.74\\
Russian	& 0.66 & 7.70 & 85.00 \\
Navajo & 77.64 & 18.38 & 26.40 \\
Maltese & 48.81 & 11.05 & 98.74\\
Arabic & 68.52 & 37.04 & 88.24 \\
Georgian & 4.46 & 0.41 & 92.47 \\
German & 0.84 & 3.32 & 89.19 \\
Finnish & 0.02 & 12.33 & 96.16\\ 
\bottomrule
\end{tabular}
\caption{Percentage of inflected word forms that have modified each part of the lemma \cite{sigmorphon} (some words can be inflected zero or multiple times, thus sums may not add to 100\%).}
\label{tab:pat}
\end{table}
\subsection{Results and Analysis}
From the results in Tab.~\ref{tab:res}, we can glean a number of observations.
First, comparing the results of our full {\bf Semi-sup} model, we can see that for all languages except Spanish, it achieves accuracies better than the single MED system, often by a large margin.
Even compared to the MED ensembled model, our single-model system is quite competitive, achieving higher accuracies for Hungarian, Navajo, Maltese, and Arabic, as well as achieving average accuracies that are state-of-the-art.

Next, comparing the different varieties of our proposed models, we can see that the semi-supervised model consistently outperforms the bidirectional model for all languages.
And similarly, the bidirectional model consistently outperforms the single direction model. 
From these results, we can conclude that the unlabeled data is beneficial to learn useful latent variables that can be used to decode the corresponding word. 

Examining the linguistic characteristics of the models in which our model performs well provides even more interesting insights.
\newcite{sigmorphon} estimate how often the inflection process involves prefix changes, stem-internal changes or suffix changes, the results of which are shown in Tab.~\ref{tab:pat}.
Among the many languages, the inflection processes of Arabic, Maltese and Navajo are relatively diverse, and contain a large amount of all three forms of inflection.
\begin{figure}[tb]
  \centering
  \includegraphics[scale=0.32]{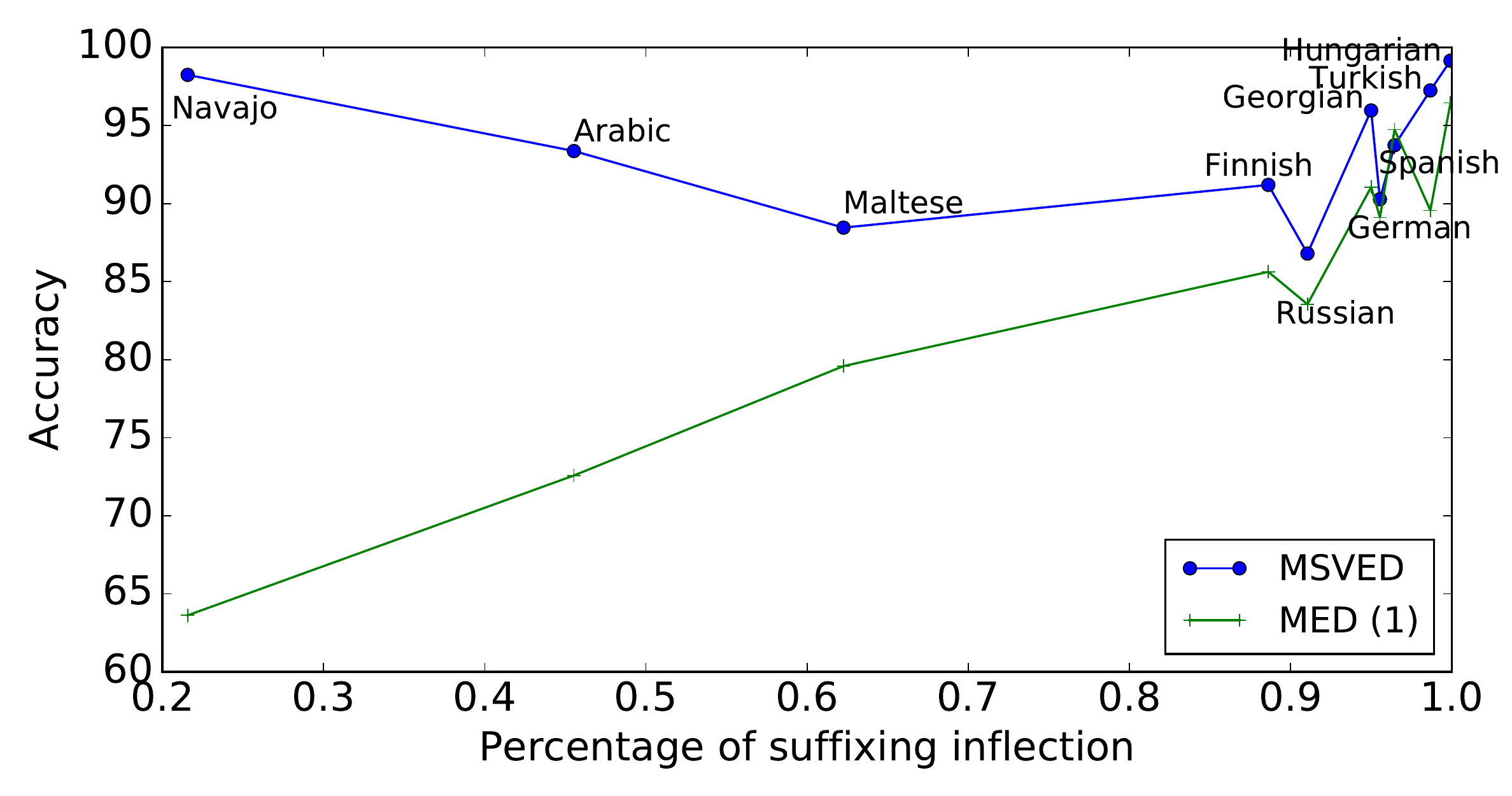}
  \vspace{-7mm}
  \caption{Performance on test data w.r.t.~the percentage of suffixing inflection. Points with the same x-axis value correspond to the same language results.}
  \label{fig:acc-morph}
  \vspace{-4mm}
\end{figure}
By examining the experimental results together with the morphological inflection process of different languages, we found that among all the languages, Navajo, Maltese and Arabic obtain the largest gains in performance compared with the ensembled MED system.
To demonstrate this visually, in Fig.~\ref{fig:acc-morph}, we compare the semi-supervised MSVED with the MED single model w.r.t.~the percentage of suffixing inflection of each language, showing this clear trend.

This strongly demonstrates that our model is agnostic to different morphological inflection forms whereas the conventional encoder-decoder with attention on the source input tends to perform better on suffixing-oriented morphological inflection.
We hypothesize that for languages that the inflection mostly comes from suffixing, transduction is relatively easy because the source and target words share the same prefix and the decoder can copy the prefix of the source word via attention.
However, for languages in which different inflections of a lemma go through different morphological processes, the inflected word and the target word may differ greatly and thus it is crucial to first analyze the lemma of the inflected word before generating the corresponding the reinflection form based on the target labels.
This is precisely what our model does by extracting the lemma representation $\zv$ learned by the variational inference model.
\begin{figure}[t!]
\captionsetup[subfigure]{labelformat=empty}
\centering
\subfloat{\includegraphics[scale=0.2]{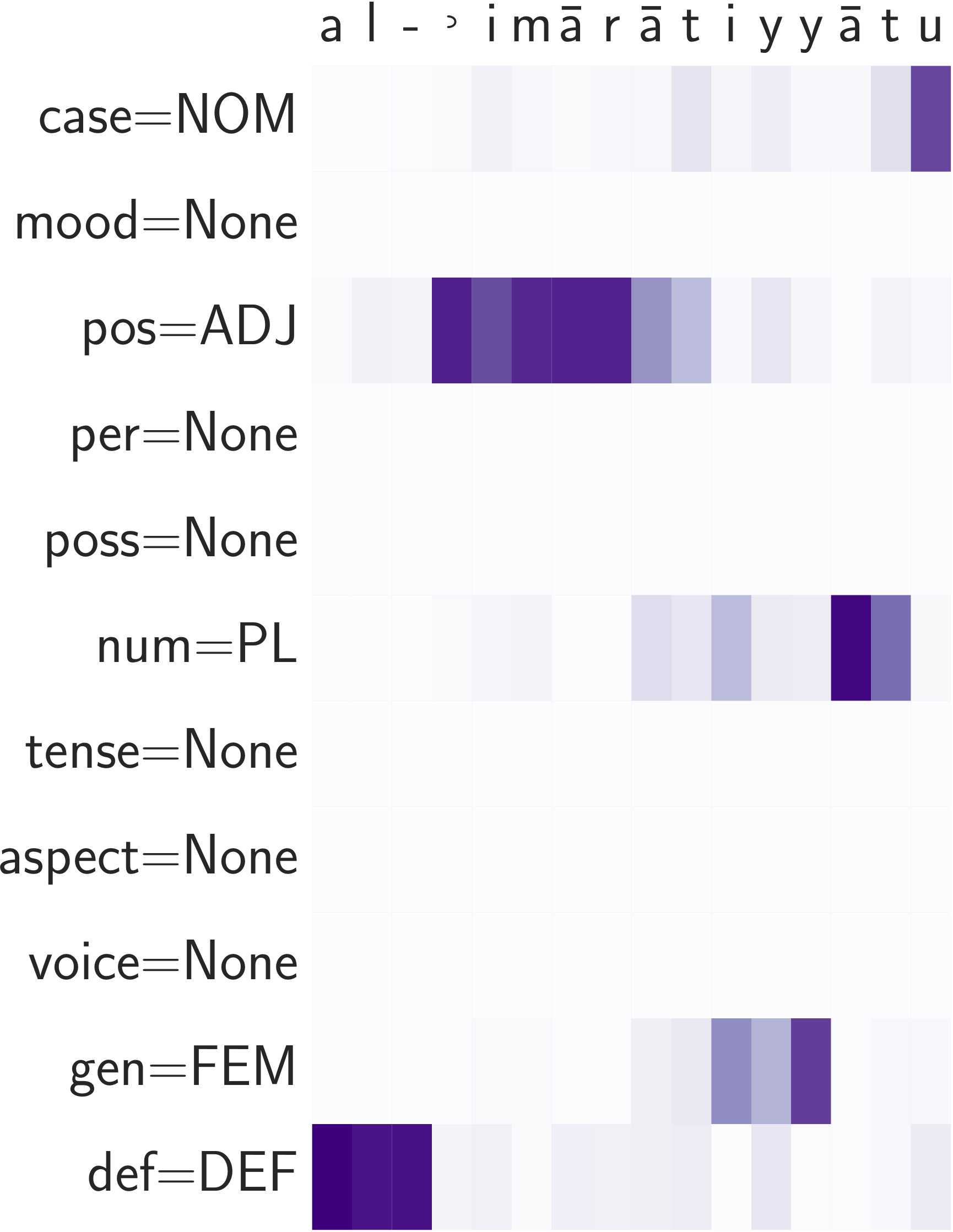}}~~~
\subfloat{\includegraphics[scale=0.2]{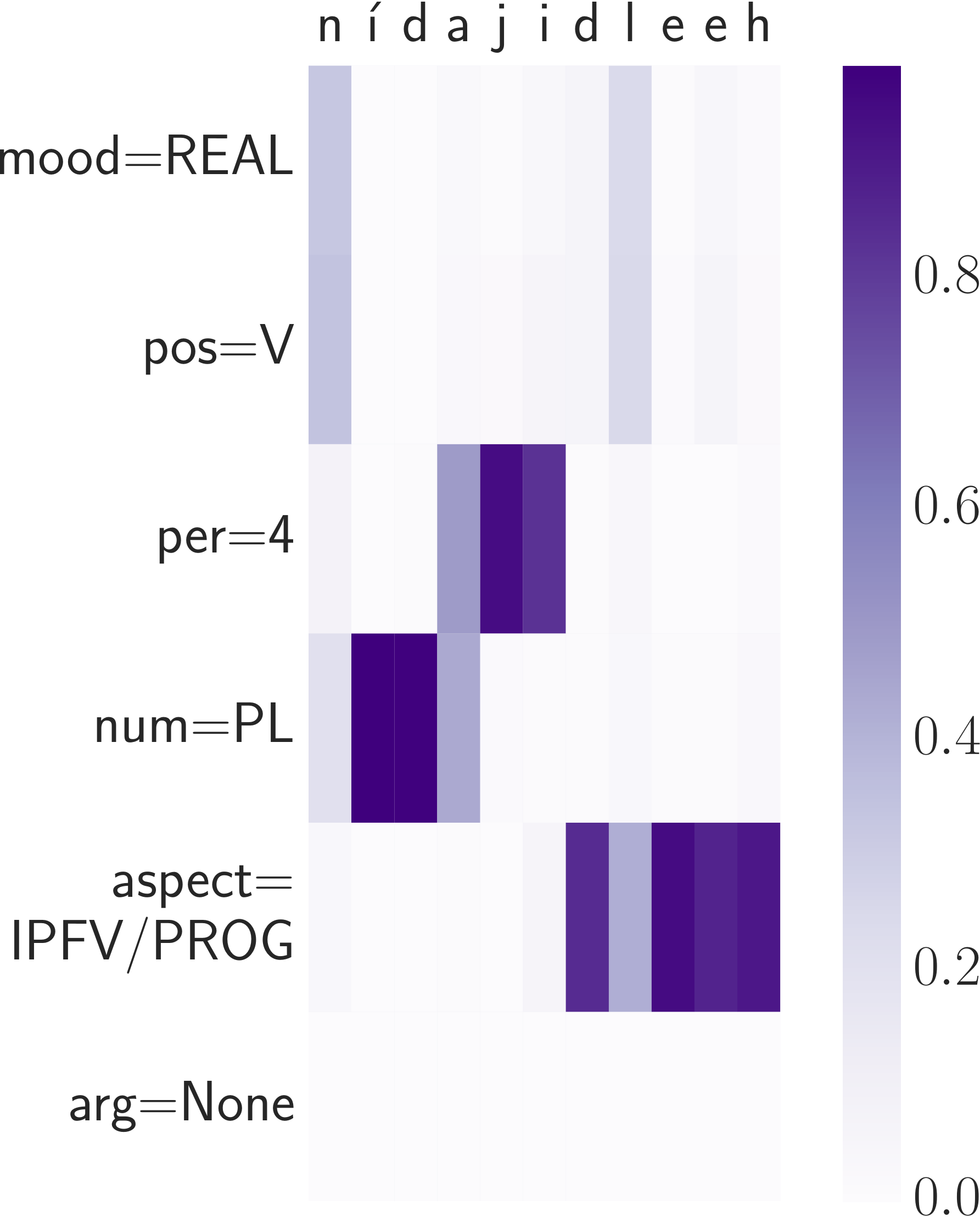}}
\caption{Two examples of attention weights on target linguistic labels: Arabic ({\bf Left}) and Navajo ({\bf Right}). When a tag equals {\tt None}, it means the word does not have this tag.}
\vspace{-2mm}
\label{fig:att}
\end{figure}
\subsection{Analysis on Tag Attention}
To analyze how the decoder attends to the linguistic labels associated with the target word, we randomly pick two words from the Arabic and Navajo test set and plot the attention weight in Fig.~\ref{fig:att}.
The Arabic word ``al-'im\={a}r\={a}tiyy\={a}tu'' is an adjective which means ``Emirati'', and its source word in the test data is ``'im\={a}r\={a}tiyyin'' \footnote{\url{https://en.wiktionary.org/wiki/\%D8\%A5\%D9\%85\%D8\%A7\%D8\%B1\%D8\%A7\%D8\%AA\%D9\%8A}}.
Both of these are declensions of ``'im\={a}r\={a}tiyy''. The source word is singular, masculine, genitive and indefinite, while the required inflection is plural, feminine, nominative and definite. We can see from the left heat map that the attention weights are turned on at several positions of the word when generating corresponding inflections. For example, ``al-'' in Arabic is the definite article that marks definite nouns.
The same phenomenon can also be observed in the Navajo example, as well as other languages, but due to space limitation, we don't provide detailed analysis here.

\subsection{Visualization of Latent Lemmas}
To investigate the learned latent representations, in this section we visualize the $\zv$ vectors, examining whether the latent space groups together words with the same lemma.
Each sample in SIGMORPHON 2016 contains source word and target words which share the same lemma. 
We run a heuristic process to assign pairs of words to groups that likely share a lemma by grouping together word pairs for which at least one of the words in each pair shares a surface form.
This process is not error free -- errors may occur in the case where multiple lemmas share the same surface form -- but in general the groupings will generally reflect lemmas except in these rare erroneous cases, so we dub each of these groups a \textit{pseudo-lemma}.

\begin{figure}[t!]
\centering
\includegraphics[scale=0.3]{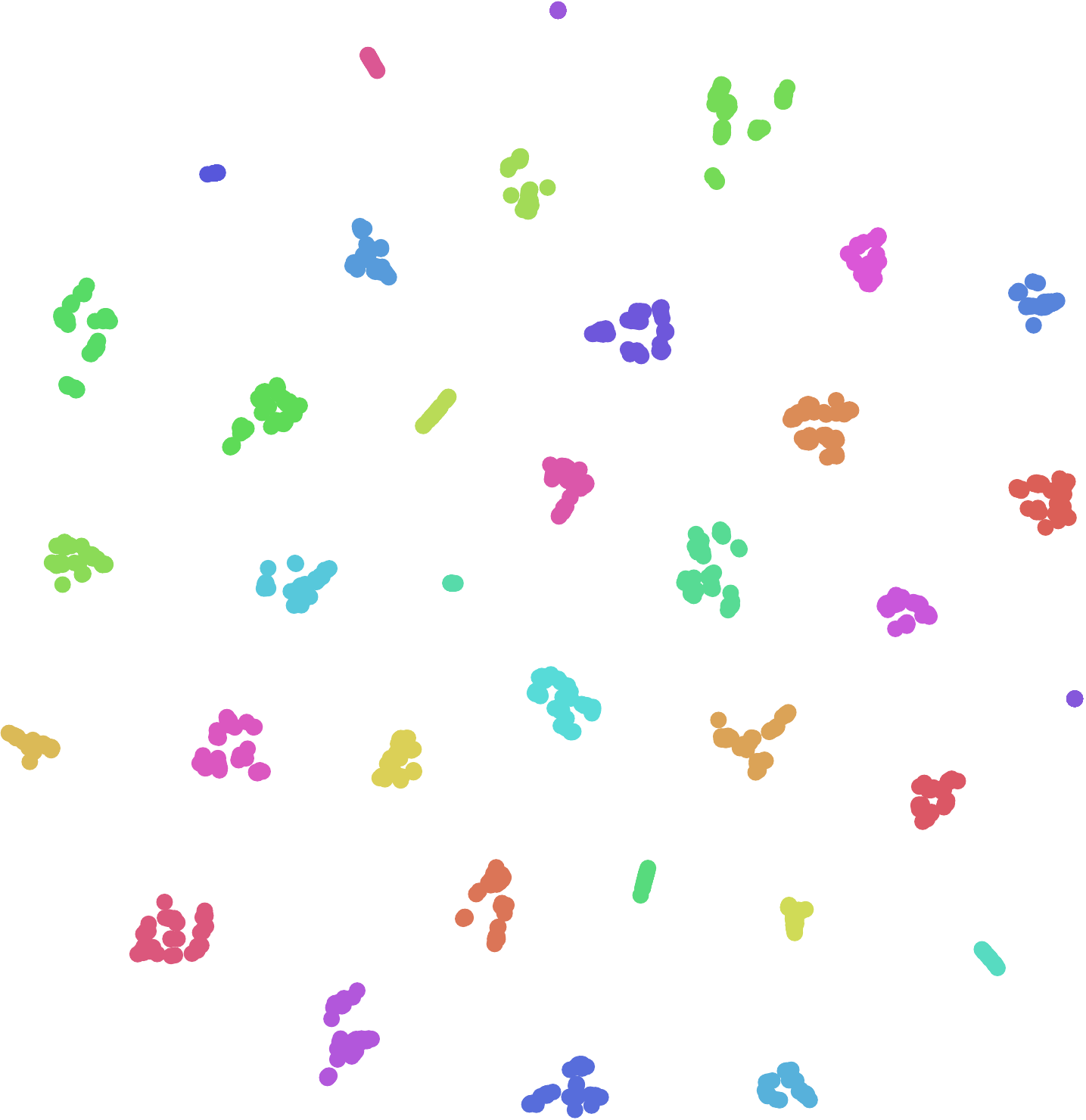}
\caption{Visualization of latent variables $\zv$ for Maltese with 35 pseudo-lemma groups in the figure.}
\vspace{-4mm}
\label{fig:tsne}
\end{figure}

In Fig.~\ref{fig:tsne}, we randomly pick 1500 words from Maltese and visualize the continuous latent vectors of these words.
We compute the latent vectors as $\mu_{\phi}(\xv)$ in the variational posterior inference (Eq. \ref{infer}) without adding the variance.
As expected, words that belong to the same pseudo-lemma (in the same color) are projected into adjacent points in the two-dimensional space.
This demonstrates that the continuous latent variable captures the canonical form of a set of words and demonstrates the effectiveness of the proposed representation.
\begin{table*}[ht!]
\centering
\small
\begin{tabular}{llcrrr}
\toprule \
\textbf{Language} & \textbf{Src Word} & \textbf{Tgt Labels} & \textbf{Gold Tgt} & \textbf{MED} & \textbf{Ours} \\\midrule
\multirow{3}{*}{Turkish} & kocama&pos=N,poss=PSS1S,case=ESS,num=SG	&kocamda &	kocama	& kocamda\\
&yaratmamdan	&pos=N,case=NOM,num=SG	
&yaratma	&yaratma	&yaratman\\
& bitimizde	&pos=N,tense=PST,per=1,num=SG	&bittik&	bitiydik	&bittim\\\midrule
\multirow{3}{*}{Maltese} & ndammhomli &	pos=V,polar=NEG,tense=PST,num=SG&	tindammhiex	& ndammejthiex &	tindammhiex\\
&tqo\.{z}\.{z}hieli &	pos=V,polar=NEG,tense=PST,num=SG&	tqo\.{z}\.{z}x	&tqo\.{z}\.{z}x &	qa\.{z}\.{z}ejtx\\
&tissikkmuhomli &	pos=V,polar=POS,tense=PST,num=PL&ssikkmulna&	tissikkmulna	&tissikkmulna\\\midrule
\multirow{3}{*}{Finnish} &	verovapaatta	&pos=ADJ,case=PRT,num=PL &	verovapaita	&verovappaita &	verovapaita\\
&turrumme	&pos=V,mood=POT,tense=PRS,num=PL	&turtunemme
&turtunemme	&turrunemme\\
&sukunimin	&pos=N,case=PRIV,num=PL &	sukunimittä	&sukunimeitta&	sukunimeitta\\
\bottomrule
\end{tabular}
\caption{Randomly picked output examples on the test data. Within each block, the first, second and third lines are outputs that ours is correct and MED's is wrong, ours is wrong and MED's is correct, both are wrong respectively.}
\label{tab:case_study}
\end{table*}
\subsection{Analyzing Effects of Size of Unlabeled Data}
\begin{figure}[tb]
  \centering
  \includegraphics[scale=0.32]{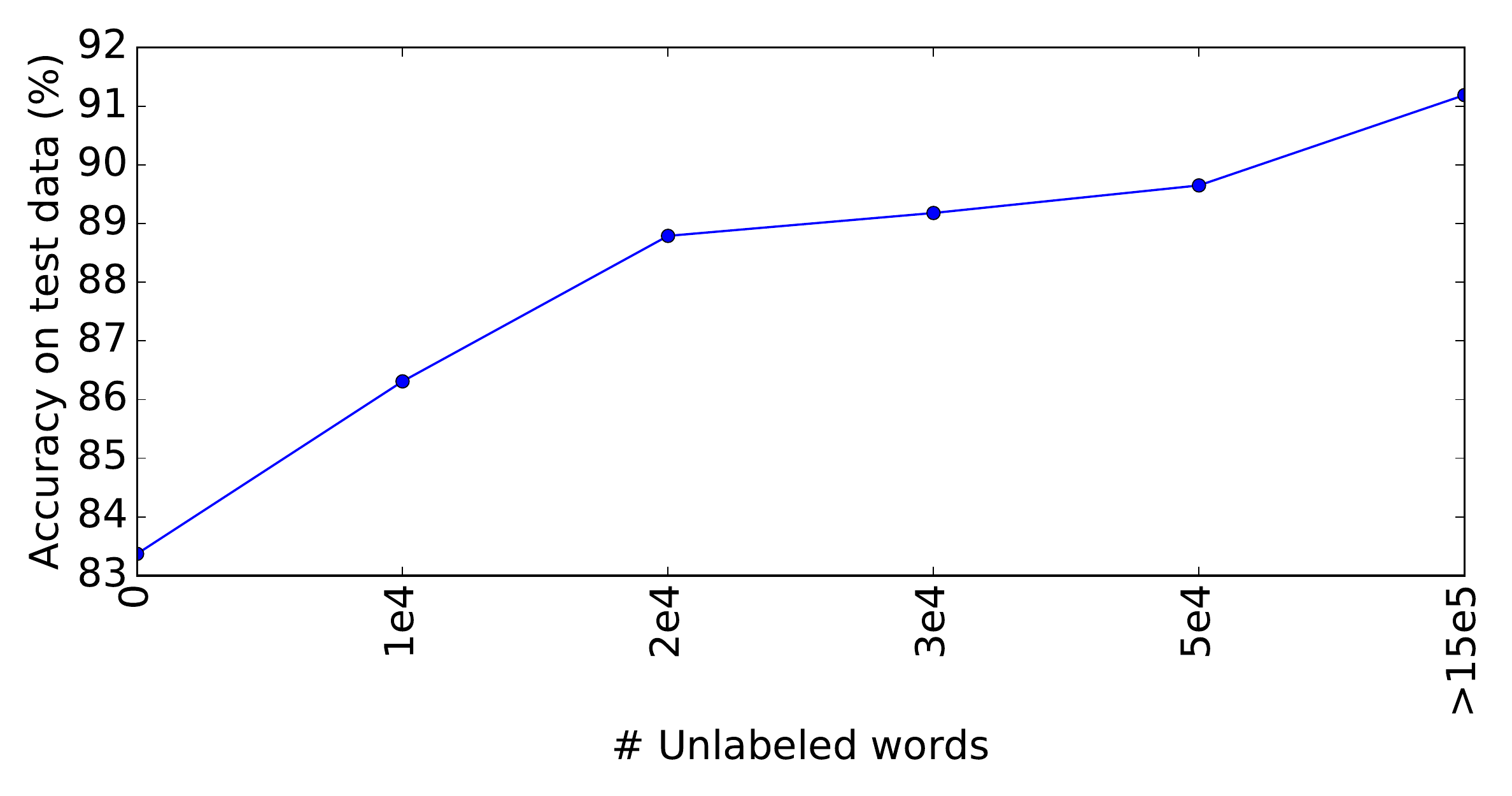}
  \vspace{-7mm}
  \caption{Performance on the German test data w.r.t.~the amount of unlabeled Wikipedia data.}
  \label{fig:ul}
  \vspace{-2mm}
\end{figure}

From Tab.~\ref{tab:res}, we can see that semi-supervised learning always performs better than supervised learning without unlabeled data.
In this section, we investigate to what extent the size of unlabeled data can help with performance.
We process a German corpus from a 2017 Wikipedia dump and obtain more than 100,000 German words. These words are ranked in order of occurrence frequency in Wikipedia.
The data contains a certain amount of noise since we did not apply any special processing.
We shuffle all unlabeled data from both the Wikipedia and the data provided in the shared task used in previous experiments, and increase the number of unlabeled words used in learning by 10,000 each time, and finally use all the unlabeled data (more than 150,000 words) to train the model.
Fig.~\ref{fig:ul} shows that the performance on the test data improves as the amount of unlabeled data increases, which implies that the unsupervised learning continues to help improve the model's ability to model the latent lemma representation even as we scale to a noisy, real, and relatively large-scale dataset.
Note that the growth rate of the performance grows slower as more data is added, because although the number of unlabeled data is increasing, the model has seen most word patterns in a relatively small vocabulary.
\subsection{Case Study on Reinflected Words}
In Tab. \ref{tab:case_study}, we examine some model outputs on the test data from the MED system and our model respectively. 
It can be seen that most errors of MED and our models can be ascribed to either over-copy or under-copy of characters.
In particular, from the complete outputs we observe that our model tends to be more aggressive in its changes, resulting in it performing more complicated transformations, both successfully (such as Maltese ``ndammhomli'' to ``tindammhiex'') and unsuccessfully (``tqo\.{z}\.{z}x'' to ``qa\.{z}\.{z}ejtx'').
In contrast, the attentional encoder-decoder model is more conservative in its changes, likely because it is less effective in learning an abstracted representation for the lemma, and instead copies characters directly from the input.

\section{Conclusion and Future Work}
In this work, we propose a multi-space variational encoder-decoder framework for labeled sequence transduction problem.
The MSVED performs well in the task of morphological reinflection, outperforming the state of the art, and further improving with the addition of external unlabeled data.
Future work will adapt this framework to other sequence transduction scenarios such as machine translation, dialogue generation, question answering, where continuous and discrete latent variables can be abstracted to guide sequence generation.

\section*{Acknowledgments}
The authors thank Jiatao Gu, Xuezhe Ma, Zihang Dai and Pengcheng Yin for their helpful discussions.
This work has been supported in part by an Amazon Academic Research Award.

\clearpage
\bibliography{morph.bbl}

\begin{thebibliography}{}
\expandafter\ifx\csname natexlab\endcsname\relax\def\natexlab#1{#1}\fi

\bibitem[{Alegria and Etxeberria(2016)}]{ehu}
I{\~n}aki Alegria and Izaskun Etxeberria. 2016.
\newblock Ehu at the sigmorphon 2016 shared task. a simple proposal:
  Grapheme-to-phoneme for inflection.
\newblock {\em In Proceedings of the 2016 Meeting of SIGMORPHON\/} .

\bibitem[{Bahdanau et~al.(2015)Bahdanau, Cho, and Bengio}]{bahdanau2014neural}
Dzmitry Bahdanau, Kyunghyun Cho, and Yoshua Bengio. 2015.
\newblock Neural machine translation by jointly learning to align and
  translate.
\newblock {\em The International Conference on Learning Representations\/} .

\bibitem[{Bayer and Osendorfer(2014)}]{bayer2014learning}
Justin Bayer and Christian Osendorfer. 2014.
\newblock Learning stochastic recurrent networks.
\newblock {\em arXiv preprint arXiv:1411.7610\/} .

\bibitem[{Bowman et~al.(2016)Bowman, Vilnis, Vinyals, Dai, Jozefowicz, and
  Bengio}]{bowman}
Samuel~R Bowman, Luke Vilnis, Oriol Vinyals, Andrew~M Dai, Rafal Jozefowicz,
  and Samy Bengio. 2016.
\newblock Generating sentences from a continuous space.
\newblock {\em Proceedings of CoNLL\/} .

\bibitem[{Chahuneau et~al.(2013)Chahuneau, Schlinger, Smith, and
  Dyer}]{chahuneau2013translating}
Victor Chahuneau, Eva Schlinger, Noah~A Smith, and Chris Dyer. 2013.
\newblock Translating into morphologically rich languages with synthetic
  phrases.
\newblock Association for Computational Linguistics.

\bibitem[{Chung et~al.(2015)Chung, Kastner, Dinh, Goel, Courville, and
  Bengio}]{vrnn}
Junyoung Chung, Kyle Kastner, Laurent Dinh, Kratarth Goel, Aaron~C Courville,
  and Yoshua Bengio. 2015.
\newblock A recurrent latent variable model for sequential data.
\newblock In {\em Advances in neural information processing systems\/}. pages
  2980--2988.

\bibitem[{Cotterell et~al.(2016)Cotterell, Kirov, Sylak-Glassman, Yarowsky,
  Eisner, and Hulden}]{sigmorphon}
R.~Cotterell, C.~Kirov, J.~Sylak-Glassman, D.~Yarowsky, J.~Eisner, and
  M.~Hulden. 2016.
\newblock The sigmorphon 2016 shared task—morphological reinflection.
\newblock In {\em Proceedings of the 54th Annual Meeting of the Association for
  Computational Linguistics\/}.

\bibitem[{Darwish and Oard(2007)}]{darwish2007adapting}
Kareem Darwish and Douglas~W Oard. 2007.
\newblock Adapting morphology for arabic information retrieval.
\newblock In {\em Arabic Computational Morphology\/}, Springer, pages 245--262.

\bibitem[{Doersch(2016)}]{doersch2016tutorial}
Carl Doersch. 2016.
\newblock Tutorial on variational autoencoders.
\newblock {\em arXiv preprint arXiv:1606.05908\/} .

\bibitem[{Durrett and DeNero(2013)}]{durrett2013supervisedmorphology}
Greg Durrett and John DeNero. 2013.
\newblock Supervised learning of complete morphological paradigms.
\newblock In {\em Proceedings of the 2013 Conference of the North American
  Chapter of the Association for Computational Linguistics: Human Language
  Technologies\/}. Association for Computational Linguistics, Atlanta, Georgia,
  pages 1185--1195.

\bibitem[{Fabius and van Amersfoort(2014)}]{fabius2014variational}
Otto Fabius and Joost~R van Amersfoort. 2014.
\newblock Variational recurrent auto-encoders.
\newblock {\em arXiv preprint arXiv:1412.6581\/} .

\bibitem[{Faruqui et~al.(2016)Faruqui, Tsvetkov, Neubig, and
  Dyer}]{faruqui2016morphological}
Manaal Faruqui, Yulia Tsvetkov, Graham Neubig, and Chris Dyer. 2016.
\newblock Morphological inflection generation using character sequence to
  sequence learning.
\newblock In {\em Proceedings of the 2016 Conference of the North American
  Chapter of the Association for Computational Linguistics: Human Language
  Technologies\/}. Association for Computational Linguistics, San Diego,
  California, pages 634--643.

\bibitem[{Gumbel and Lieblein(1954)}]{gumbel1954statistical}
Emil~Julius Gumbel and Julius Lieblein. 1954.
\newblock Statistical theory of extreme values and some practical applications:
  a series of lectures.
\newblock {\em US Government Printing Office Washington\/} .

\bibitem[{J. et~al.(2017)J., Mnih, and Teh}]{gumbel2}
Maddison~Chris J., Andriy Mnih, and Yee~Whye Teh. 2017.
\newblock The concrete distribution: A continuous relaxation of discrete random
  variables.
\newblock In {\em The International Conference on Learning Representations.\/}.

\bibitem[{Jang et~al.(2017)Jang, Gu, and Poole.}]{gumbel}
Eric Jang, Shixiang Gu, and Ben Poole. 2017.
\newblock Categorical reparameterization with gumbel-softmax.
\newblock In {\em The International Conference on Learning Representations.\/}.

\bibitem[{Johnson et~al.(2016)Johnson, Schuster, Le, Krikun, Wu, Chen, Thorat,
  Vi{\'e}gas, Wattenberg, Corrado et~al.}]{johnson2016google}
Melvin Johnson, Mike Schuster, Quoc~V Le, Maxim Krikun, Yonghui Wu, Zhifeng
  Chen, Nikhil Thorat, Fernanda Vi{\'e}gas, Martin Wattenberg, Greg Corrado,
  et~al. 2016.
\newblock Google's multilingual neural machine translation system: Enabling
  zero-shot translation.
\newblock {\em arXiv preprint arXiv:1611.04558\/} .

\bibitem[{Kann et~al.(2016)Kann, Cotterell, and Sch{\"u}tze}]{kann2016neural}
Katharina Kann, Ryan Cotterell, and Hinrich Sch{\"u}tze. 2016.
\newblock Neural multi-source morphological reinflection.
\newblock {\em arXiv preprint arXiv:1612.06027\/} .

\bibitem[{Kann and Sch\"{u}tze(2016{\natexlab{a}})}]{med}
Katharina Kann and Hinrich Sch\"{u}tze. 2016{\natexlab{a}}.
\newblock Med: The lmu system for the sigmorphon 2016 shared task on
  morphological reinflection.
\newblock In {\em In Proceedings of the 14th SIGMORPHON Workshop on
  Computational Research in Phonetics, Phonology, and Morphology\/}. Berlin,
  Germany.

\bibitem[{Kann and Sch\"{u}tze(2016{\natexlab{b}})}]{med2}
Katharina Kann and Hinrich Sch\"{u}tze. 2016{\natexlab{b}}.
\newblock Single-model encoder-decoder with explicit morphological
  representation for reinflection.
\newblock In {\em In Proceedings of the 54th Annual Meeting of the Association
  for Computational Linguistics\/}. Berlin, Germany.

\bibitem[{Kikuchi et~al.(2016)Kikuchi, Neubig, Sasano, Takamura, and
  Okumura}]{kikuchi2016outputlength}
Yuta Kikuchi, Graham Neubig, Ryohei Sasano, Hiroya Takamura, and Manabu
  Okumura. 2016.
\newblock Controlling output length in neural encoder-decoders.
\newblock In {\em Proceedings of the 2016 Conference on Empirical Methods in
  Natural Language Processing\/}. Association for Computational Linguistics,
  Austin, Texas, pages 1328--1338.

\bibitem[{Kingma et~al.(2014)Kingma, Mohamed, Rezende, and Welling}]{semivae}
Diederik~P Kingma, Shakir Mohamed, Danilo~Jimenez Rezende, and Max Welling.
  2014.
\newblock Semi-supervised learning with deep generative models.
\newblock In {\em Advances in Neural Information Processing Systems\/}.
  Montr\'{e}al, Canada, pages 3581--3589.

\bibitem[{Kingma and Welling(2014)}]{vae}
D.P. Kingma and M.~Welling. 2014.
\newblock Auto-encoding variational bayes.
\newblock In {\em The International Conference on Learning Representations\/}.

\bibitem[{Ko{\v{c}}isk{\`y} et~al.(2016)Ko{\v{c}}isk{\`y}, Melis, Grefenstette,
  Dyer, Ling, Blunsom, and Hermann}]{kovcisky2016semantic}
Tom{\'a}{\v{s}} Ko{\v{c}}isk{\`y}, G{\'a}bor Melis, Edward Grefenstette, Chris
  Dyer, Wang Ling, Phil Blunsom, and Karl~Moritz Hermann. 2016.
\newblock Semantic parsing with semi-supervised sequential autoencoders.
\newblock {\em the 2016 Conference on Empirical Methods in Natural Language
  Processing (EMNLP)\/} .

\bibitem[{Maal{\o}e et~al.(2016)Maal{\o}e, S{\o}nderby, S{\o}nderby, and
  Winther}]{maaloe2016auxiliary}
Lars Maal{\o}e, Casper~Kaae S{\o}nderby, S{\o}ren~Kaae S{\o}nderby, and Ole
  Winther. 2016.
\newblock Auxiliary deep generative models.
\newblock {\em Proceedings of the 33rd International Conference on Machine
  Learning\/} .

\bibitem[{Maddison et~al.(2014)Maddison, Tarlow, and
  Minka}]{maddison2014sampling}
Chris~J Maddison, Daniel Tarlow, and Tom Minka. 2014.
\newblock A* sampling.
\newblock In {\em Advances in Neural Information Processing Systems\/}. pages
  3086--3094.

\bibitem[{Miao and Blunsom(2016)}]{miao2016language}
Yishu Miao and Phil Blunsom. 2016.
\newblock Language as a latent variable: Discrete generative models for
  sentence compression.
\newblock {\em the 2016 Conference on Empirical Methods in Natural Language
  Processing (EMNLP)\/} .

\bibitem[{Nicolai et~al.(2016)Nicolai, Hauer, Arnaud, and Kondrak}]{alberta}
Garrett Nicolai, Bradley Hauer, Adam~St. Arnaud, and Grzegorz Kondrak. 2016.
\newblock Morphological reinflection via discriminative string transduction.
\newblock {\em In Proceedings of the 2016 Meeting of SIGMORPHON\/} .

\bibitem[{Ostling(2016)}]{ostling2016morphological}
Robert Ostling. 2016.
\newblock Morphological reinflection with convolutional neural networks.
\newblock {\em In Proceedings of the 14th SIGMORPHON Workshop on Computational
  Research in Phonetics, Phonology, and Morphology\/} page~23.

\bibitem[{Sennrich et~al.(2016)Sennrich, Haddow, and
  Birch}]{sennrich2016controlling}
Rico Sennrich, Barry Haddow, and Alexandra Birch. 2016.
\newblock Controlling politeness in neural machine translation via side
  constraints.
\newblock In {\em Proceedings of the 2016 Conference of The North American
  Chapter of the Association for Computational Linguistics (NAACL)\/}. pages
  35--40.

\bibitem[{Taji et~al.(2016)Taji, Eskander, Habash, and Rambow}]{taji}
Dima Taji, Ramy Eskander, Nizar Habash, and Owen Rambow. 2016.
\newblock The columbia university - new york university abu dhabi sigmorphon
  2016 morphological reinflection shared task submission.
\newblock {\em In Proceedings of the 2016 Meeting of SIGMORPHON\/} .

\bibitem[{Tokuda et~al.(2002)Tokuda, Masuko, Miyazaki, and
  Kobayashi}]{tokuda2002multi}
Keiichi Tokuda, Takashi Masuko, Noboru Miyazaki, and Takao Kobayashi. 2002.
\newblock Multi-space probability distribution hmm.
\newblock {\em IEICE TRANSACTIONS on Information and Systems\/} 85(3):455--464.

\bibitem[{Toutanova et~al.(2008)Toutanova, Suzuki, and
  Ruopp}]{toutanova2008applying}
Kristina Toutanova, Hisami Suzuki, and Achim Ruopp. 2008.
\newblock Applying morphology generation models to machine translation.
\newblock In {\em Proceedings of the 46th Annual Meeting of the Association for
  Computational Linguistics\/}. pages 514--522.

\bibitem[{Zeiler(2012)}]{zeiler2012adadelta}
Matthew~D Zeiler. 2012.
\newblock Adadelta: an adaptive learning rate method.
\newblock {\em arXiv preprint arXiv:1212.5701\/} .

\bibitem[{Zhang et~al.(2016)Zhang, Xiong, Su, Duan, and Zhang}]{vnmt}
Biao Zhang, Deyi Xiong, Jinsong Su, Hong Duan, and Min Zhang. 2016.
\newblock Variational neural machine translation.
\newblock {\em Proceedings of the 54th Annual Meeting of the Association for
  Computational Linguistics\/} .

\end{thebibliography}
\bibliographystyle{acl_natbib}


\end{document}